\documentclass[journal=jacsat,manuscript=article]{achemso}

\usepackage[version=3]{mhchem} 
\usepackage[usenames, dvipsnames]{color}
\usepackage{amssymb}
\usepackage{algorithm2e}
\usepackage{titlesec}
\usepackage{url}
\setkeys{acs}{articletitle=true}
\setkeys{acs}{chaptertitle=true}

\newcommand{\eqn}[2]{\begin{equation}#1\label{#2}\end{equation}}
\DeclareMathOperator*{\argmin}{arg\,min}

\newcommand*{\blauw}[1]{#1}
\newcommand*{\groen}[1]{#1}

\author{Andrew W. Long}
\affiliation
{Department of Materials Science and Engineering, University of Illinois at Urbana-Champaign, 1304 W Green St, Urbana, IL, USA}
\author{Andrew L. Ferguson}
\affiliation
{Department of Materials Science and Engineering, University of Illinois at Urbana-Champaign, 1304 W Green St, Urbana, IL, USA}
\alsoaffiliation
{Department of Chemical and Biomolecular Engineering, University of Illinois at Urbana-Champaign, 600 South Mathews Avenue, Urbana, IL, USA}
\email{alf@illinois.edu}
\phone{(217) 300-2354}
\fax{(217) 333-2736}
\title[Landmark DMaps] {Landmark Diffusion Maps (L-dMaps): Accelerated manifold learning out-of-sample extension}

\begin{document}

\clearpage
\newpage




\begin{abstract}
\noindent Diffusion maps are a nonlinear manifold learning technique based on harmonic analysis of a diffusion process over the data. Out-of-sample extensions with computational complexity $\mathcal{O}(N)$, where $N$ is the number of points comprising the manifold, frustrate applications to online learning applications requiring rapid embedding of high-dimensional data streams. We propose landmark diffusion maps (L-dMaps) to reduce the complexity to $\mathcal{O}(M)$, where $M \ll N$ is the number of landmark points selected using pruned spanning trees or k-medoids. Offering $(N/M)$ speedups in out-of-sample extension, L-dMaps enables the application of diffusion maps to high-volume and/or high-velocity streaming data. We illustrate our approach on three datasets: the Swiss roll, molecular simulations of a \ce{C24H50} polymer chain, and biomolecular simulations of alanine dipeptide. We demonstrate up to 50-fold speedups in out-of-sample extension for the molecular systems with less than 4\% errors in manifold reconstruction fidelity relative to calculations over the full dataset.
\end{abstract}

\clearpage
\newpage

\noindent \textbf{\Large{Keywords}}

\noindent diffusion maps; harmonic analysis; spectral graph theory; nonlinear dimensionality reduction; molecular simulation

\vspace{10mm}

\noindent \textbf{\Large{Abbreviations}}

\noindent dMaps -- diffusion maps; L-dMaps -- landmark diffusion maps; PCA - principal component analysis; L-Isomap - landmark Isomap; PST -- pruned spanning tree; RMS -- root mean squared; RMSD -- root mean squared deviation

\clearpage
\newpage

\noindent \textbf{\Large{Highlights}}

\begin{itemize}
\item Landmark diffusion maps (L-dMaps) applies diffusion maps over a subset of data points
\item L-dMaps offers orders of magnitude speedups in out-of-sample embedding of new data
\item These accelerations enable nonlinear embedding of high-velocity streaming data
\end{itemize}

\clearpage
\newpage

\section{Introduction}
Linear and nonlinear dimensionality reduction algorithms have found broad applications in diverse application domains including computer vision\cite{Cho2010_CompVision}, recommendation engines\cite{Sarwar2000_WebKDD}, anomaly detection\cite{Patcha2007_CompNet}, and protein folding\cite{Das2006_PNAS}. These techniques all seek to discover within high-dimensional datasets low-dimensional projections that preserve the important collective variables governing the global system behavior and to which the remaining degrees of freedom are effectively slaved\cite{Transtrum2015_JChemPhys,Machta2013_Science,Ferguson2010_PNAS,Zwanzig2001_NEQSM,Coifman2008_MultiModSim}. Efficient and robust identification of low-dimensional descriptions is of value in improved time-series forecasting\cite{Pena2006_DimRed}, understanding protein folding pathways\cite{Ferguson2010_BiophysJ}, and generating automated product recommendations\cite{Linden2003_InternetComp}. 

Linear techniques such as principal component analysis (\groen{PCA})\cite{PCA_Springer}, multidimensional scaling\cite{Borg2005_Springer}, and random projection\cite{Bingham2001_KDD}, have proven successful due to their simplicity and relative efficiency in identifying these low-dimensional manifolds. Nonlinear methods, such as Isomap\cite{Tenenbaum2000_Science}, LLE\cite{Roweis2000_Science}, and diffusion maps\cite{Coifman2006_AppCompHarmAnal}, have proven successful in areas with more complex manifolds where linear techniques fail, trading the simplicity and efficiency of linear methods for the capability to discover more complex nonlinear relationships and provide more parsimonious representations. Of these methodologies, diffusion maps have received substantial theoretical and applied interest for their firm mathematical grounding in the harmonic analysis of diffusive modes over the high-dimensional data\cite{Coifman2005_PNAS,Coifman2006_AppCompHarmAnal}. Specifically, under relatively mild assumptions on the data, the collective order parameters spanning the low-dimensional nonlinear projection discovered by diffusion maps correspond to the slowest dynamical modes of the relaxation of a probability distribution evolving under a random walk over the data\cite{Coifman2005_PNAS,Coifman2006_AppCompHarmAnal,Nadler2006_NIPS,Ferguson2011_ChemPhysLett}. This endows the low-dimensional embedding discovered by diffusion maps with two attractive properties. First, ``diffusion distances'' in the high-dimensional space measuring the accessibility of one system configuration from another map to Euclidean distances in the low-dimensional projection \cite{Coifman2005_PNAS,Nadler2006_NIPS,Ferguson2011_ChemPhysLett}. Second, the collective order parameters supporting the low-dimensional embedding correspond to the important nonlinear collective modes containing the most variance in the data \cite{Ferguson2011_ChemPhysLett}. These properties of the diffusion map have been exploited in diverse applications, including the understanding of protein folding and dynamics\cite{Ferguson2010_BiophysJ,Mansbach2015_JChemPhys}, control of self-assembling Janus colloids\cite{Long2015_SoftMatter}, tomographic image reconstruction\cite{Coifman2008_IEEE}, image completion\cite{Gepshtein2013_TIP}, graph matching\cite{Hu2016_IDA}, and automated tissue classification in pathology slices\cite{Coifman2005_PNAS}.

Practical application of diffusion maps typically involves two distinct but related operations: (i) analysis of a high-dimensional dataset in $\mathcal{R}^K$ to discover and define a low-dimensional projection in $\mathcal{R}^k$ where $k < K$ and (ii) projection of new out-of-sample points into the low-dimensional manifold. Discovery of the manifold requires calculation of pairwise distances between all data points to construct a Markov matrix over the high-dimensional data, and subsequent diagonalization of this matrix to perform a spectral decomposition of the corresponding random walk \cite{Nadler2006_NIPS,Pan1999_Eigen,Hu2016_IDA}. This eigendecomposition yields an ordered series of increasingly faster relaxing modes of the random walk, and the identification of a gap in the eigenvalue spectrum of implied time scales informs the effective dimensionality of the underlying low-dimensional manifold and collective coordinates with which to parametrize it \cite{Coifman2005_PNAS,Coifman2006_AppCompHarmAnal,Nadler2006_NIPS,Coifman2008_MultiModSim,Belkin2003_NeurComp,Ferguson2011_ChemPhysLett}. Exact calculation of all pairwise distances is of complexity $\mathcal{O}(N^2)$, although it is possible to exploit the exponential decay in the hopping probabilities to threshold these matrix dements to zero and to avoid calculation of all distance pairs using clustering or divide and conquer approaches\cite{Kao2008_Algo}. Diagonalization of the $N$-by-$N$ matrix has complexity $\mathcal{O}(N^3)$\cite{Golub2013_Matrix}, but the typically sparse nature of the Markov matrix admits sparse Lanczos algorithms that reduce the complexity to $\mathcal{O}(N^2 + N E_{nz})$, where $E_{nz}$ is the number of non-zero matrix elements \cite{Hu2016_IDA,Bechtold2006_Springer,Larsen1998_Lanczos}. Calculation of only the top $l \ll N$ eigenvectors can further reduce the cost to $\mathcal{O}(l N + l E_{nz})$ \cite{Bechtold2006_Springer,Hu2016_IDA}. The overall complexity of the discovery of the nonlinear low-dimensional projection is then $\mathcal{O}(N^2 + l N + l E_{nz}) \sim \mathcal{O}(N^2)$. 


Projection of out-of-sample points into the nonlinear manifold is complicated by the unavailability of an explicit transformation matrix for the low-dimensional projection \cite{Ferguson2011_ChemPhysLett}. Na\"{i}vely, one may augment the original data with the $N_\mathrm{new}$ samples and diagonalize the augmented system, which is an exact but exceedingly expensive operation to perform for each new sample. Accordingly, a number of approximate interpolation techniques have been proposed\cite{Bengio2004_NIPS}, including principal component analysis-based approaches\cite{Aizenbud_PCA_OOSE}, Laplacian pyramids\cite{Rabin2012_ICDM}, and the Nystr\"{o}m extension\cite{Fowlkes2004_TPAMI,Lafon2006_TPAMI,Baker1977_Numerical}. The Nystr\"{o}m extension is perhaps the simplest and most widely used and scales as $\mathcal{O}(N)$, requiring the calculation of pairwise distances with the $N$ points constituting the low-dimensional manifold. Linear scaling with the size of the original dataset can be prohibitively costly for online dimensionality reduction applications to high-velocity and/or high-volume streaming data, where fast and efficient embedding of new data points is of paramount importance. For example, in online threat or anomaly detection where excursions of the system into unfavorable regions of the nonlinear manifold must be quickly recognized in order to take corrective action\cite{Eskin2002_Anomaly}, in robotic motion planning where movements are generated based on localized paths through a low-dimensional manifold to maintain kinematic admissibility while navigating obstacles\cite{Mahoney2010_ICRA,Chen2016_IEEE}, and in parallel replica dynamics\cite{Voter1998_PhysRevB} or forward flux sampling simulations\cite{Allen2009_JPhysCondMat,Escobedo2009_JPhysCondMat} of biomolecular folding where departures from a particular region of the manifold can be used to robustly and rapidly identify conformational changes in any one of the simulation replicas and permit responsive reinitialization of the replicas to make maximally efficient use of computational resources.

To reduce the computational complexity of the application of diffusion maps to streaming data we propose a controlled approximation based on the identification of a subset of $M \ll N$ ``landmark'' data points with which to construct the original manifold and embed new data. Reducing the number of points participating in these operations can offer substantial computational savings, and the degree of approximation can be controlled and tuned by the number and location of the landmarks. Our approach is inspired by and analogous to the landmark Isomap (\groen{L-Isomap}) adaptation of the original Isomap nonlinear manifold learning approach due to Tenenbaum, de Silva, and Langford \cite{Tenenbaum2000_Science,deSilva2002_NIPS}, which has demonstrated robust landscape recovery and computational savings considering small numbers of randomly selected landmark points. Silva \textit{et al.}\ subsequently introduced a systematic means to select landmarks based on $L_1$-regularized minimization of a least-squares objective function \cite{Silva2005_NIPS}. As observed by de Silva and Tenenbaum \cite{deSilva2002_NIPS}, the underlying principle of this approach is analogous to global positioning using local distances, whereby a point can be uniquely located within the manifold given sufficient distance measurements to points distributed over its surface \cite{Singer2008_PNAS}. In analogy with this work, we term our approach the landmark diffusion maps (\groen{L-dMaps}). 

It is the purpose of this manuscript to introduce L-dMaps, in which the selection of $M \ll N$ landmark points reduces the computational complexity of  the out-of-sample projection of a new data point from $\mathcal{O}(N)$ to $\mathcal{O}(M)$ offering speedups $S \propto N/M$, which can be a substantial factor when the landmarks constitute a small fraction of the data points constituting the manifold. The use of landmarks also substantially reduces the memory requirements, leading to savings in both CPU and RAM requirements that enable applications of diffusion maps to higher volume and velocity streaming data than is currently possible. 

The structure of this paper is as follows. In \blauw{Materials and Methods}, we introduce the computational and algorithmic details of our L-dMaps approach along with theoretical error bounds on its fidelity relative to diffusion maps applied to the full dataset.  In \blauw{Results and Discussion}, we demonstrate and analyze the accuracy and performance of L-dMaps on three test systems -- the Swiss roll, molecular simulations of a \ce{C24H50} polymer chain, and biomolecular simulations of alanine dipeptide -- in which we report up to 50-fold speedups in out-of-sample extension with less than 4\% errors in manifold reconstruction fidelity relative to those calculated by dMaps applied to the full dataset. In our \blauw{Conclusions} we close with an appraisal of our approach and its applications, and an outlook for future work.

\section{Materials and Methods}

First, we briefly describe the original diffusion maps (\groen{dMaps}) approach and the Nystr\"{o}m extension for out-of-sample projection. Second, we introduce landmark diffusion maps (\groen{L-dMaps}) presenting two algorithms for systematic identification of landmark points -- one fully automated spanning tree approach, and one based on k-medoids that can be tuned to achieve specific error tolerances -- and the subsequent use of these landmarks to perform nonlinear manifold discovery and Nystr\"{o}m projection of new data. Third, we develop theoretical estimates of L-dMaps error bounds based on a first-order perturbation expansion in the errors introduced by the use of landmarks compared to consideration of the full dataset. Finally, we detail the three datasets upon which we demonstrate and validate our approach: the Swiss roll, molecular simulations of a \ce{C24H50} polymer chain, and biomolecular simulations of alanine dipeptide.

\subsection{Diffusion map dimensionality reduction technique} \label{sec:dmap}

The diffusion map is a nonlinear dimensionality reduction technique that discovers low-dimensional manifolds within high-dimensional datasets by performing harmonic analysis of a random walk constructed over the data to identify nonlinear collective variables containing the predominance of the variance in the data\cite{Coifman2006_AppCompHarmAnal,Coifman2005_PNAS,Coifman2008_MultiModSim,Nadler2006_NIPS}. The first step in applying diffusion maps is to compute a measure of similarity between the $N$ high-dimensional data points to construct the $N$-by-$N$ pairwise distance matrix $\mathbf{d}$ with elements,
\eqn{d_{ij}=||\vec{x}_i-\vec{x}_j||, }{eqn:dmap_pair}
where $||.||$ is an appropriate distance metric for the system under consideration (e.g. Euclidean, Hamming, earth movers distance, rotationally and translational aligned root mean squared deviation (\groen{RMSD})). These distances are then used to define a random walk over the data by defining hopping probabilities $A_{ij}$ from point $i$ to point $j$ as proportional to the convolution of $d_{ij}$ with a Gaussian kernel,
\eqn{A_{ij}=\exp \left(-\frac{d_{ij}^2}{2\epsilon} \right), }{eqn:dmap_kernel}
where $\epsilon$ is a soft-thresholding bandwidth that limits transitions between points within an $\sqrt{\epsilon}$ neighborhood. Systematic procedures exist to select appropriate values of $\epsilon$ for a particular dataset.\cite{Coifman2008_IEEE,Ferguson2010_PNAS} Forming the diagonal matrix $\mathbf{D}$ containing the row sums of the $\mathbf{A}$ matrix, 
\eqn{D_{ii}=\sum_{j=1}^N A_{ij},}{eqn:dmap_norm}
we normalize the hopping probabilities to obtain the Markov matrix $\mathbf{M}$,
\eqn{\mathbf{M} = \mathbf{D}^{-1}\mathbf{A}, }{eqn:dmap_Markov}
describing a discrete diffusion process over the data. Although other choices of kernels are possible, the symmetric Gaussian kernel is the infinitesimal generator of a diffusion process such that the Markov matrix is related to the normalized graph Laplacian,
\eqn{\mathbf{L} = \mathbf{I}-\mathbf{M}, }{eqn:dmap_Laplacian}
where $\mathbf{I}$ is the identity matrix, and in the limit of $N \rightarrow \infty$ and $\epsilon \rightarrow 0$ the matrix $\mathbf{L}$ converges to a Fokker-Planck operator describing a continuous diffusion process over the high-dimensional data \cite{Coifman2006_AppCompHarmAnal,Coifman2005_PNAS,Nadler2006_AppCompHarmAnal,Ferguson2010_PNAS}.

Diagonalizing $\mathbf{M}$ by solving the $N$-by-$N$ eigenvalue problem,
\begin{align}
\mathbf{M} \mathbf{\Psi} &= \mathbf{\Psi} \mathbf{\Lambda} \label{eqn:dmap_dmap}
\end{align}
where $\mathbf{\Lambda}$ is a diagonal matrix holding the eigenvalues $\{\lambda_i\}_{i=1}^N$ in non-ascending order and $\mathbf{\Psi}=\{\vec{\psi}_i\}_{i=1}^N$ is a matrix of right column eigenvectors corresponding to the relaxation modes and implied time scales of the random walk \cite{Coifman2006_AppCompHarmAnal,Coifman2005_PNAS}. By the Markov property the top pair $\{ \lambda_1=1, \vec{\psi}_1 = \vec{1} \}$ are trivial, and the steady state probability distribution over the high-dimensional point cloud given by the top left eigenvector $\vec{\phi}_1 = \mathrm{diag}(\mathbf{D})$ \cite{Ferguson2010_PNAS}. The graph Laplacian $\mathbf{L}$ and Markov matrix $\mathbf{M}$ share left $\mathbf{\Phi}$ and right $\mathbf{\Psi}$ biorthogonal eigenvectors, and the eigenvectors of $\mathbf{L}$ are related to those of $\mathbf{M}$ as $\lambda_i^L = 1 - \lambda_i$ \cite{Ferguson2010_PNAS}. Accordingly, the leading eigenvectors of $\mathbf{M}$ are the slowest relaxing modes of the diffusion process described by the graph Laplacian $\mathbf{L}$ \cite{Coifman2005_PNAS}.

A gap in the eigenvalue spectrum exposes a separation of time scales between slow and fast relaxation modes, informing an embedding into the slow collective modes of the discrete diffusion process that excludes the remaining quickly relaxing fast degrees of freedom \cite{Nadler2006_AppCompHarmAnal}. Identifying this gap at $\lambda_{k+1}$ informs an embedding into the top $k$ non-trivial eigenvectors,
\eqn{\vec{x}_i\rightarrow\{\vec{\psi}_2(i),\vec{\psi}_3(i),\ldots,\vec{\psi}_{k+1}(i)\}.}{eqn:microstate_map}
This diffusion mapping can be interpreted as a nonlinear projection of the high-dimensional data in $\mathcal{R}^K$ onto a low-dimensional ``intrinsic manifold'' in $\mathcal{R}^k$ discovered within the data where $k < K$. The diffusion map determines both the dimensionality $k$ of the intrinsic manifold and good collective order parameters $\{\vec{\psi}_l\}_{l=2}^{k+1}$ with which to parameterize it. As detailed above, the overall complexity of manifold discovery and projection via diffusion maps is $\mathcal{O}(N^2)$, where efficient diagonalization routines leave this calculation dominated by calculation of pairwise distances.

\subsubsection{Nystr\"{o}m extension for out-of-sample points} \label{sec:nystrom}

The Nystr\"{o}m extension presents a means to embed new data points outside the original $N$ used to define the low-dimensional nonlinear embedding by approximate interpolation of the new data point within the low-dimensional manifold\cite{Fowlkes2004_TPAMI,Lafon2006_TPAMI,Baker1977_Numerical,Sonday2009_PhysRevE,Bengio2004_NIPS}. This operation proceeds by computing the distances of the new point to the $N$ existing points defining the manifold $d_{\textrm{new},j}=||\vec{x}_\textrm{new} - \vec{x}_j||$, and using these values to compute $A_{\textrm{new},j}=\exp \left(-\frac{d_{\textrm{new},j}^2}{2\epsilon} \right)$ and augment the $\mathbf{M}$ matrix with an additional row $M_{\textrm{new},j} = \left(\sum_{j=1}^N A_{\textrm{new},j} \right)^{-1} A_{\textrm{new},j}$ \cite{Bengio2004_NIPS,Sonday2009_PhysRevE}. The projected coordinates of the new point onto the $k$-dimensional intrinsic manifold defined by the top $l = 2 \ldots (k+1)$ non-trivial eigenvectors of $\mathbf{M}$ are then given by,
\eqn{\vec{\psi}_{l}(\textrm{new})=\frac{1}{\lambda_l}\sum_{j=1}^N  M_{\textrm{new},j}\vec{\psi}_l(j).}{eqn:nystrom}
The computational complexity for the Nystr\"{o}m projection of a single new data point is $\mathcal{O}(N)$, requiring $N$ distance calculations in the original $K$-dimensional space and then projection onto the $k$-dimensional manifold. Projections are exact for points in the original dataset, accurate for the interpolation of new points within the kernel bandwidth $\sqrt{\epsilon}$ of the intrinsic manifold defined by the $N$ original points, but poor for extrapolations to points residing beyond this distance away from the manifold \cite{Bengio2004_NIPS,Ferguson2011_ChemPhysLett,Ferguson2011_JChemPhys}.

\subsection{Landmark diffusion maps (L-dMaps)}

The application of diffusion maps to the online embedding of streaming data is limited by the $\mathcal{O}(N)$ computational complexity of the out-of-sample extension that requires the calculation of pairwise distances of each new data point with the $N$ points defining the embedding. Here we propose landmark diffusion maps (\groen{L-dMaps}) that employs a subset of $M \ll N$ points providing adequate support for discovery and construction of the low-dimensional embedding to reduce this complexity to $\mathcal{O}(M)$. In this way, we trade-off errors in the embedding of new data with projection speedups that scale in inverse proportion to the number of landmarks $M$. We quantify the reconstruction error introduced by the landmarking procedure, and demonstrate that these errors can be made arbitrarily small by selection of sufficiently many landmark points. The use of landmarks has previously been demonstrated in the L-Isomap variant of the Isomap dimensionality reduction technique to offer substantial gains in computational efficiency\cite{deSilva2002_NIPS}.
 

\subsubsection{Landmark selection}

Landmarking can be conceived as a form of lossy compression that represents localized groups of points in the high-dimensional feature space by attributing them to a central representative landmark point. Provided the landmarks are sufficiently well distributed over the intrinsic manifold mapped out by the data in the high-dimensional space, then the pairwise distances between landmarks to a new out-of-sample data point provide sufficient distance constraints to accurately embed the new point onto the manifold\cite{deSilva2002_NIPS,Singer2008_PNAS}. The original L-Isomap algorithm proposed landmarks be selected randomly\cite{deSilva2002_NIPS}, and a subsequent sophistication by Silva \textit{et al.}\ proposed a selection procedure based on $L_1$-regularized minimization of a least-squares objective function \cite{Silva2005_NIPS}. In this work, we propose two efficient and systematic approaches to selection: a pruned spanning tree (\groen{PST}) approach that offers an automated means to select landmarks, and a k-medoids approach that allows the user to tune the number of landmarks to trade-off speed and embedding fidelity to achieve a particular error tolerance. Both approaches require pre-computation of the $N$-by-$N$ pairwise distances matrix, making them expensive for large datasets\cite{Silva2005_NIPS}. However, it is the primary goal of this work to select good landmarks for the rapid and efficient embedding of streaming data into an existing manifold, so the high one-time overhead associated with their selection is of secondary importance relative to optimal landmark identification for subsequent online performance.

\textbf{Pruned spanning tree (PST) landmark selection.} The square root of the soft-thresholding bandwidth $\sqrt{\epsilon}$ employed by diffusion maps defines the characteristic step size of the random walk over the high dimensional data (\blauw{Eqn.\ \ref{eqn:dmap_kernel}})\cite{Ferguson2010_PNAS}. In order to reliably construct a low-dimensional embedding, the graph formed by applying this neighborhood threshold to the pairwise distance matrix must be fully connected to assure than no point is unreachable from any other (i.e., the Markov matrix $\mathbf{M}$ is irreducible) \cite{Ferguson2011_ChemPhysLett}. This assures that a single connected random walk can be formed over the data, and that the diffusion map will discover a single unified intrinsic manifold as opposed to a series of disconnected manifolds each containing locally connected subsets of the data. This connectivity criterion imposes two requirements on the selection of landmarks $\{\vec{z}_i\}_{i=1}^M \in \mathcal{R}^K$ as a subset of the data points $\{\vec{x}_i\}_{i=1}^N \in \mathcal{R}^K$: (i) all $N$ points are within a $\sqrt{\epsilon}$ neighborhood of (i.e., covered by) a landmark, 
\eqn{\forall \vec{x}_i, \exists \vec{z}_{j} \mid ||\vec{x}_i-\vec{z}_j|| \le \sqrt{\epsilon},}{eqn:covering} 
and (ii) the graph of pairwise distances over the landmarks is fully connected, with each landmark point within a distance of $\sqrt{\epsilon}$ of at least one other,
\eqn{\forall \vec{z}_i, \exists \vec{z}_{j\neq i} \mid ||\vec{z}_i-\vec{z}_j|| \le \sqrt{\epsilon}.}{eqn:connectivity} 
In practice, a threshold of a few multiples of $\sqrt{\epsilon}$ may be sufficient to maintain coverage and connectivity.

These coverage and connectivity conditions motivate a landmark selection procedure based on spanning trees of the pairwise distances matrix $\mathbf{d}$ that naturally enforces both of these constraints and identifies landmarks that ensure out-of-sample points residing within the manifold can be embedded within a neighborhood $\sqrt{\epsilon}$ of a landmark point. Residing within the characteristic step size of the random walk, this condition is expected to produce accurate interpolative out-of-sample extensions using the Nystr\"{o}m extension. As described above, extrapolative extensions are expected to perform poorly for distances greater than $\sqrt{\epsilon}$\cite{Bengio2004_NIPS,Ferguson2011_ChemPhysLett,Ferguson2011_JChemPhys}. First, we form the binary adjacency matrix $\mathbf{G}$ by hard-thresholding the $N$-by-$N$ pairwise distances matrix $\mathbf{d}$ (\blauw{Eqn.\ \ref{eqn:dmap_pair}}),
\eqn{
G_{ij}=
\begin{cases}
1 & \mbox{if } d_{ij} \le \sqrt{\epsilon}, i\ne j \\
0 & \mbox{otherwise}
\end{cases}.
}{eqn:adjacency}
The binary adjacency matrix defines a new graph in which two data points $\vec{x}_i$ and $\vec{x}_j$ are connected if and only if $G_{ij}=1$. Next, we seek the minimal subset of edges that contains no cycles and connects all nodes in the graph, which is equivalent to identifying a spanning tree representation of the graph $\mathbf{T}$ that may be determined in many ways\cite{Cormen2009_Algo}. We elect to use Prim's algorithm\cite{Prim1957_Bell}, which randomly selects a root node then recursively adds edges between tree nodes and unassigned nodes until all nodes are incorporated into the tree. As the edge weights of $\mathbf{G}$ are either 0 or 1, Prim's algorithm at each step randomly selects an edge from $\mathbf{G}$ connecting an unassigned node to a node in $\mathbf{T}$. By only creating new connections between tree nodes and non-tree nodes, this method guarantees that $\mathbf{T}$ is cycle-free and, provided that $\mathbf{G}$ is connected, is a spanning tree of $\mathbf{G}$. Finally, we recognize that each leaf node lies within $\sqrt{\epsilon}$ of their parent, meaning that all leaves of the tree can be pruned, with the remaining nodes comprising a pruned spanning tree (\groen{PST}) defining a set of landmarks $\{\vec{z}_i\}$ satisfying both the covering (\blauw{Eqn.\ \ref{eqn:covering}}) and connectivity (\blauw{Eqn.\ \ref{eqn:connectivity}}) conditions. We summarize PST landmark identification procedure in \blauw{Algorithm \ref{alg:PST}}.

\RestyleAlgo{boxruled}
\begin{algorithm}[ht]
  \caption{PST landmark selection\label{alg:PST}}
  \textbf{Input:} $\{\vec{x}_i\}_{i=1}^N$, $\mathbf{G}=(\mathbf{V_G},\mathbf{E_G})$\\
  Initialize tree $\mathbf{T}=(\mathbf{V_T},\mathbf{E_T})$ by selecting a random node $i$, $\mathbf{V_T}=\{i\},\mathbf{E_T}=\emptyset$ \\
  Construct the spanning tree:\\
  while $\mathbf{V_T} \neq \mathbf{V_G}$ do \\
  \quad Gather set of all edges $\mathbf{E_N}$ between tree and unassigned nodes:\\
  \quad\quad\quad $\mathbf{E_N}=\{uv:u\in \mathbf{V_T}, v\in \mathbf{V_G\backslash V_T},uv\in \mathbf{E_G}\}$\\
  \quad Randomly add edge $mn\in \mathbf{E_N}$ to $\mathbf{T}$:\\
  \quad\quad\quad $\mathbf{V_T}=\mathbf{V_T}\cup\{n\},\mathbf{E_T}=\mathbf{E_T}\cup\{mn\}$\\
  end while\\
  Identify leaf nodes $\mathbf{V_L}$ of $\mathbf{T}$ (nodes of degree 1 in $\mathbf{T}$):\\
  \quad $\mathbf{V_L}=\{u:u\in\mathbf{V_T},\mathrm{deg}(u)=1\}$ \\
  Select all non-leaf nodes: \\
  \quad $ \{\vec{z}\}=\{\vec{x}_i:i\in \mathbf{V_T}\backslash\mathbf{V_L}\}$\\
  
 \textbf{Output:} $\{\vec{z}\}$ $\equiv$ landmarks
\end{algorithm}

\textbf{K-medoid landmark selection.} Growing and pruning a spanning tree offers an automated means to identify landmarks that ensure any out-of-sample point can be interpolatively embedded within a $\sqrt{\epsilon}$ neighborhood of a landmark. This procedure is expected to offer good reconstruction accuracy, but the error tolerance is not directly controlled by the user. Accordingly, we introduce a second approach to landmark selection that allows the user to tune the number of landmark points to trade-off computational efficiency against embedding accuracy in the Nystr\"{o}m extension to achieve a particular runtime target or error tolerance relative to the use of all $N$ data points. Specifically, we partition the data into a set of $M$ distinct clusters, and define landmarks within each of these clusters to achieve pseudo-optimal coverage of the intrinsic manifold for a particular number of landmark points. Numerous partitioning techniques are available, including spectral clustering\cite{Von2007_StatComp}, affinity propagation\cite{Frey2007_Science}, and agglomerative hierarchical clustering\cite{Day1984_JClass}. We use the well-known k-medoids algorithm, using Voronoi iteration to update and select medoid points\cite{Park2009_ExpSysApp}. Compared to k-means clustering, k-medoids possesses the useful attribute that cluster prototypes are selected as medoids within the data points constituting the cluster as opposed to as linear combinations defining the cluster mean. We select the initial set of landmarks randomly from the pool of all samples, although we note that alternate seeding methods exist such as $k$-means++\cite{Arthur2007_SIAM}. We summarize the k-medoids landmark selection in \blauw{Algorithm \ref{alg:kmed}}.

\RestyleAlgo{boxruled}
\begin{algorithm}[ht]
  \caption{K-medoid landmark selection\label{alg:kmed}}
  \textbf{Input:} $\{\vec{x}_i\}_{i=1}^N$, $M$, maxItr\\
  Randomly select initial landmarks $\{\vec{z}^{(0)}_j\}_{j=1}^M \subseteq \{\vec{x}_i\}_{i=1}^N$ \\
  t = 0 \\
  do \\
  \quad Assign points to clusters: \\
  \quad\quad\quad$S_i^{(t)}=\lbrace \vec{x}_j : ||\vec{x}_j-\vec{z}_i^{(t)}|| \leq ||\vec{x}_j-\vec{z}_m^{(t)}|| \textrm{ for all } m=1,\ldots,M \rbrace$ \\

  \quad Update cluster medoid: \\
  \quad\quad\quad$\vec{z}_i^{(t+1)} = \argmin_{\vec{x}_m\in S_i^{(t)}} \sum_{\vec{x}_j\in S_i^{(t)}} ||\vec{x}_m-\vec{x}_j||$ \\
  \quad t = t+1 \\
  while $\{\vec{z}^{(t)}\}\neq \{\vec{z}^{(t-1)}\}$) and (t $<$ maxItr)\\
  \textbf{Output:} $\{\vec{z}^{(t)}\}$ $\equiv$ landmarks
\end{algorithm}

\subsubsection{Landmark intrinsic manifold discovery}\label{sec:landmark_dmap}

The primary purpose of landmark identification is to define an ensemble of $M$ supports for the efficient out-of-sample extension projection of streaming data. The nonlinear manifold can be defined by applying diffusion maps to all $N$ data points. The expensive $\mathcal{O}(N^2)$ calculation of the pairwise distances matrix $\mathbf{d}$ has already been performed for the purposes of landmark identification, leaving only a relatively cheap $\mathcal{O}(l N + l E_{nz})$ calculation of the top $l$ eigenvectors where $E_{nz}$ is the number of non-zero elements in the matrix \cite{Bechtold2006_Springer,Hu2016_IDA,Larsen1998_Lanczos}. Nevertheless, having identified these landmarks, additional computational savings may be achieved by constructing the manifold using only the $M$ landmarks.

Na\"{i}ve application of diffusion maps to the $M$ landmarks will yield a poor reconstruction of the original manifold since these landmark points do not preserve the density distribution of the $N$ original points over the high-dimensional feature space. Ferguson \emph{et al.}\ previously proposed a means to efficiently apply diffusion maps to datasets containing multiple copies of each point in the context of recovering nonlinear manifolds from biased data\cite{Ferguson2011_JChemPhys}. We adapt this approach to apply diffusion maps to only the landmark points while approximately maintaining the density distribution of the full dataset. Given a set of landmark points $\{\vec{z}_i\}_{i=1}^M$ we may characterize the local density of points in the high-dimensional space around each landmark by counting the number of data points residing within the Voronoi volume of each landmark point $\vec{z}_i$ defined by the set,
\eqn{S_i = \{\vec{x}_j : ||\vec{x}_j-\vec{z}_i||\leq ||\vec{x}_j-\vec{z}_m|| \textrm{ for all } m=1,\ldots,M\}. }{eqn:clustering}
Following \blauw{Ref.\ \citenum{Ferguson2011_JChemPhys}}, we now apply diffusion maps to the $M$ landmark points each weighted by a multiplicity $c_i=|S_i|$. Mathematically, this corresponds to solving the $M$-by-$M$ eigenvalue problem analogous to that in \blauw{Eqn.\ \ref{eqn:dmap_dmap}} of \blauw{Section \ref{sec:dmap}},
\begin{align}
\tilde{\mathbf{M}} \tilde{\mathbf{C}} \tilde{\mathbf{\Psi}} &= \tilde{\mathbf{\Psi}} \tilde{\mathbf{\Lambda}}, \label{eqn:block_eig}
\end{align}
where $\tilde{\mathbf{M}} = \tilde{\mathbf{D}}^{-1} \tilde{\mathbf{A}}$, the elements of $\tilde{\mathbf{A}}$ are given by,
\eqn{\tilde{A}_{ij}=\exp \left(-\frac{||\vec{z}_i-\vec{z}_j||^2}{2\epsilon} \right), }{eqn:ldmap_kernel}
defining the unnormalized hopping probability between landmark points $\vec{z}_i$ and $\vec{z}_j$, $\tilde{\mathbf{C}}$ is a diagonal matrix containing the multiplicity of each landmark point,
\eqn{\tilde{C}_{ii}=c_i=|S_i|,}{eqn:ldmap_mult}
$\tilde{\mathbf{D}}$ is a diagonal matrix with elements,
\eqn{\tilde{D}_{ii}=\sum_{j=1}^M \tilde{A}_{ij} \tilde{C}_{jj},}{eqn:ldmap_norm}
and $\tilde{\mathbf{\Lambda}}$ is a diagonal matrix holding the eigenvalues $\{\tilde{\lambda}_i\}_{i=1}^M$ in non-ascending order, and $\tilde{\mathbf{\Psi}}=\{\vec{\tilde{\psi}}_i\}_{i=1}^M$ is the matrix of right column eigenvectors. It can be shown that by enforcing the normalization condition $\tilde{\mathbf{C}} \vec{\tilde{\psi}}_i \cdot \vec{\tilde{\psi}}_i = 1$ on the eigenvectors, that the diffusion map embedding,
\eqn{\vec{z}_i\rightarrow\{\vec{\tilde{\psi}}_2(i),\vec{\tilde{\psi}}_3(i),\ldots,\vec{\tilde{\psi}}_{k+1}(i)\},}{eqn:ldmap_microstate_map}
is precisely that which would be obtained from applying diffusion maps to an ensemble of points in which each landmark point $\vec{z}_i$ is replicated $c_i$ times, and the $M$ non-zero eigenvalues $\{\lambda_i\}_{i=1}^M$ are identical to $\{\tilde{\lambda}_i\}_{i=1}^M$ \cite{Ferguson2011_JChemPhys}.

The net result of this procedure is that we can define the intrinsic manifold by considering only the landmark points and diagonalizing a $M$-by-$M$ matrix as opposed to a $N$-by-$N$ matrix with an attendant reduction in computational complexity from $\mathcal{O}(l N + l E_{nz})$ to $\mathcal{O}(l M + l E_{nz})$. The diffusion mapping in \blauw{Eqn.\ \ref{eqn:ldmap_microstate_map}} defines a reduced $M$-point intrinsic manifold that can be considered a lossy compression of the complete $N$-point manifold, and which can be stored in a smaller memory footprint\cite{deSilva2002_NIPS}. For large datasets, the computational and memory savings associated with calculation and storage of this manifold can be significant. Although not necessary, if desired the $(N-M)$ non-landmark points can be projected into the intrinsic manifold using the landmark Nystr\"{o}m extension \blauw{described in the next section}. This is also precisely the procedure that will be used to perform out-of-sample embeddings of new data points not contained within the original $N$ data points.

Geometrically, the approximation we make in formulating the reduced eigenvalue problem is that all points within the Voronoi cell of a landmark point are equivalent to the landmark point itself, weighting each landmark point according to the number of points inside its Voronoi volume. This is a good assumption provided that the variation between the points within each Voronoi volume is small relative to the variation over the rest of the high-dimensional space, and becomes exact in the limit that every point is treated as a landmark (i.e., $M = N$). In \blauw{Section \ref{sec:meth_err}} we place theoretical bounds on the errors introduced by this approximation in the nonlinear projection of new out-of-sample points onto the intrinsic manifold relative to that which would have been computed by explicitly considering all $N$ points using the original diffusion map.

\subsubsection{Landmark Nystr\"{o}m extension}\label{sec:landmark_nystrom}

The heart of our L-dMaps approach is the nonlinear projection of new out-of-sample points using the reduced manifold defined by the $M$ landmark points as opposed to the full $N$-point manifold. Nystr\"{o}m embeddings of new points $\vec{x}_\mathrm{new}$ over the reduced manifold proceed in an analogous manner to that detailed in \blauw{Section \ref{sec:nystrom}}, but now considering only the landmark points. Specifically, we compute the distance of the new point to all landmarks $\{\vec{z}_j\}_{j=1}^M$ to compute elements $\tilde{A}_{\textrm{new},j}=\exp\left(-\frac{||\vec{x}_\textrm{new}-\vec{z}_j||^2}{2\epsilon}\right)$ with which to augment the $\tilde{\mathbf{M}}$ matrix with an additional row with elements $\tilde{M}_{\textrm{new},j} = \left(\sum_{j=1}^M \tilde{A}_{\textrm{new},j} \tilde{C}_{jj} \right)^{-1} \tilde{A}_{\textrm{new},j}$. The projected coordinates of the new point onto the $k$-dimensional reduced intrinsic manifold defined by the top $l = 2 \ldots (k+1)$ non-trivial eigenvectors of $\tilde{\mathbf{M}}$ is,
\eqn{\vec{\tilde{\psi}}_{l}(\textrm{new})=\frac{1}{\tilde{\lambda}_j}\sum_{j=1}^M  \tilde{M}_{\textrm{new},j} \tilde{C}_{jj} \vec{\tilde{\psi}}_l(j).}{eqn:nystrom_block}
This landmark form of the Nystr\"{o}m extension reduces the computational complexity from $\mathcal{O}(N)$ to $\mathcal{O}(M)$ by reducing both the number of distance computations and the size of the matrix-vector multiplication. The attendant runtime speedup $S \propto N/M$ can be substantial for $M \ll N$, offering accelerations to permit the application of diffusion map out-of-sample extension to higher volume and higher velocity streaming data than is currently possible.

\subsection{Landmark error estimation}\label{sec:meth_err}

The diffusion mapping in \blauw{Eqn.\ \ref{eqn:ldmap_microstate_map}} defines a reduced intrinsic manifold comprising the $M \ll N$ landmark points that we subsequently use to perform efficient landmark Nystr\"{o}m projections of out-of-sample data using \blauw{Eqn.\ \ref{eqn:nystrom_block}}. As detailed in \blauw{Section \ref{sec:landmark_dmap}} the $k$ leading eigenvectors $\{\vec{\psi}_i^\prime\}_{i=2}^{k+1}$ defining the intrinsic manifold come from the solution of a $N$-by-$N$ eigenvalue problem in which each landmark point is weighted by the number of points falling in its Voronoi volume $c_i=|S_i|$, that we solve efficiently and exactly by mapping it to the reduced $M$-by-$M$ eigenvalue problem in \blauw{Eqn.\ \ref{eqn:block_eig}}\cite{Ferguson2011_JChemPhys}. The approximation we make in formulating this eigenvalue problem is to consider each point in the Voroni volume of each landmark point as identical to the landmark itself. Were we not to make the landmark approximation, we would be forced to solve a different $N$-by-$N$ eigenvalue problem explicitly treating all $N$ points using the original diffusion map with $k$ leading eigenvectors $\{\vec{\psi}_i\}_{i=2}^{k+1}$. In using $M \ll N$ landmarks we massively accelerate the out-of-sample embedding of new points, but the penalty we pay is that the resulting intrinsic manifold we discover is not exactly equivalent to that which would be discovered by the original diffusion map in the limit $M \rightarrow N$. In this section we estimate the errors in manifold reconstruction introduced by the landmarking procedure by developing approximate analytical expressions for the discrepancy between the landmark $\{\vec{\psi}_i^\prime\}_{i=2}^{k+1}$ and true $\{\vec{\psi}_i\}_{i=2}^{k+1}$ eigenvectors parameterizing the intrinsic manifold.

We formulate our approximate landmark eigenvalue problem by collapsing the set of points $S_k = \{\vec{x}_j : ||\vec{x}_j-\vec{z}_k||\leq ||\vec{x}_j-\vec{z}_m|| \textrm{ for all } m=1,\ldots,M\}$ within the Voronoi volume of each landmark point $\{\vec{z}_k\}_{k=1}^M$ onto the landmark itself. This amounts to perturbing each data point $\vec{x}_i$ in the high dimensional space by a vector $\vec{\Delta}_{i} = \vec{z}_\gamma - \vec{x}_i$ where $\vec{z}_\gamma$ is the landmark point within the Voronoi volume of which $\vec{x}_i$ falls. The elements of the $N$-by-$N$ pairwise distance matrix between are correspondingly perturbed from $d_{ij} = ||\vec{x}_i-\vec{x}_j||$ to $d_{ij}^\prime = || (\vec{x}_i + \vec{\Delta}_i) - (\vec{x}_j + \vec{\Delta}_j) ||$. This introduces perturbations $\delta_{ij} = d_{ij}^\prime - d_{ij}$ into the pairwise distances, the precise form of which depends on the choice of distance metric $||.||$. By propagating these differences through the eigenvalue problem formulated by the original diffusion map over the true (unshifted) locations of the $N$ points as a first-order perturbation expansion, we will develop analytical approximations valid in the limit of small perturbations (i.e., sufficiently many landmark points) for the corresponding perturbations in the eigenvalues and eigenvectors introduced by our landmarking procedure. We observe that for a given choice of landmark points and distance metric, the elements $\delta_{ij}$ are explicitly available, allowing is to use our terminal expressions to predict the errors introduced for a particular choice of landmarks. In \blauw{Section \ref{sec:dmap_acc}} we will validate our analytical predictions against errors calculated by explicit numerical solution of the full and landmark eigenvalue problems.

We first show how to compute the perturbations in our kernel matrix $\mathbf{\delta A}$ and diagonal row sum matrix $\mathbf{\delta D}$ as a function of the pairwise distance perturbations $\delta_{ij}$, from which we estimate perturbations in the Markov matrix $\mathbf{\delta M}$ using this expression. We then use these values to develop analytical approximations for the perturbations to the eigenvalues $\mathbf{\delta \Lambda}$ and eigenvectors $\mathbf{\delta \Psi}$ due to the landmark approximation. The perturbation analysis of the eigenvalue problem presented below follows a similar development to that developed by Deif \cite{Deif1995_JCompApplMath}.

Starting from the full diffusion map eigenvalue problem over the original data $\mathbf{M} \mathbf{\Psi} = \mathbf{\Psi} \mathbf{\Lambda}$ (\blauw{Eqn.\ \ref{eqn:dmap_dmap}}), we define the perturbed eigenvalue problem under the perturbation in the pairwise distances matrix as,
\begin{align}
&\mathbf{M}^\prime \mathbf{\Psi}^\prime = \mathbf{\Psi}^\prime \mathbf{\Lambda}^\prime \nonumber \\
\Rightarrow \; &(\mathbf{M + \delta M}) (\mathbf{\Psi + \delta \Psi}) = (\mathbf{\Psi + \delta \Psi}) (\mathbf{\Lambda + \delta \Lambda}), \label{eqn:dmap_perturb}
\end{align}
and where from \blauw{Eqn.\ \ref{eqn:dmap_Markov}},
\begin{align}
\mathbf{M + \delta M} &= (\mathbf{D + \delta D})^{-1} (\mathbf{A + \delta A}) \nonumber \\
&= (\mathbf{D}^{-1} - \mathbf{\delta D} \mathbf{D}^{-2}) (\mathbf{A + \delta A}) + \mathcal{O}(\mathbf{\delta D}^2) \nonumber \\
&= \mathbf{D}^{-1} \mathbf{A} + \mathbf{D}^{-1} \mathbf{\delta A} - \mathbf{\delta D} \mathbf{D}^{-2} \mathbf{A} + \mathcal{O}(\mathbf{\delta D}^2, \mathbf{\delta D} \mathbf{\delta A}) \nonumber \\
&= \mathbf{M} + \mathbf{D}^{-1} \mathbf{\delta A} - \mathbf{\delta D} \mathbf{D}^{-2} \mathbf{A} + \mathcal{O}(\mathbf{\delta D}^2, \mathbf{\delta D} \mathbf{\delta A}) \nonumber \\
\Rightarrow \mathbf{\delta M} &= \mathbf{M}^\prime - \mathbf{M} \nonumber \\
&= \mathbf{D}^{-1} \mathbf{\delta A} - \mathbf{\delta D} \mathbf{D}^{-2} \mathbf{A} + \mathcal{O}(\mathbf{\delta D}^2, \mathbf{\delta D} \mathbf{\delta A})\label{eqn:Merror} 
\end{align}
where in going from the first line to the second we have employed the Maclaurin expansion,
\begin{align}
\left( D_{ii} + \delta D_{ii} \right)^{-1} &= \frac{1}{D_{ii}} - \frac{\delta D_{ii}}{D_{ii}^2} + \mathcal{O}(\delta D_{ii}^2),
\end{align}
and \blauw{Eqn.\ \ref{eqn:dmap_Markov}} in going from the third to the fourth. The perturbations in the elements of the Markov matrix $\mathbf{\delta M}$ may then be estimated to first order in the perturbations as,
\begin{align}
\delta M_{ij} &= \frac{\delta A_{ij}}{D_{ii}} - \frac{\delta D_{ii}}{D_{ii}^{2}} A_{ij}. \label{eqn:Mijerror}
\end{align}
To estimate the $\mathbf{\delta A}$ and $\mathbf{\delta D}$  required by this expression as a function of the perturbations in the pairwise distances $\delta_{ij}$, we commence from the expression for the elements of $\mathbf{A}^\prime$ keeping terms in $\delta_{ij}$ to first order,
\begin{align}
A^\prime_{ij} &= \exp \left(-\frac{(d_{ij}+\delta_{ij})^2}{2\epsilon} \right) \nonumber \\
&= \exp \left(-\frac{d_{ij}^2}{2\epsilon}\right) \exp \left(-\frac{d_{ij}\delta_{ij}}{\epsilon}\right) \exp \left(-\frac{\delta_{ij}^2}{2\epsilon}\right) \nonumber \\
&= A_{ij} \left( 1 - \frac{d_{ij}\delta_{ij}}{\epsilon} \right) + \mathcal{O}(\delta_{ij}^2), \label{eqn:err1}
\end{align}
from which the perturbations in $A_{ij}$ follow as,
\begin{align}
\delta A_{ij} &= A^\prime_{ij} - A_{ij} \nonumber \\
&= \left( -\frac{d_{ij}\delta_{ij}}{\epsilon} \right) A_{ij} + \mathcal{O}(\delta_{ij}^2). \label{eqn:err3}
\end{align}
The elements of the diagonal $\mathbf{D}^\prime$ and $\mathbf{D}$ matrices are computed from the $\mathbf{A}^\prime$ and $\mathbf{A}$ row sums respectively, from which the perturbations in $D_{ii}$ follow immediately as,
\begin{align}
\delta D_{ii} &= D_{ii}^\prime - D_{ii} \nonumber \\
&= \sum_{j=1}^N A_{ij}^\prime - \sum_{j=1}^N A_{ij} \nonumber \\
&= \sum_{j=1}^N \left[ A_{ij} \left( 1 - \frac{d_{ij}\delta_{ij}}{\epsilon} \right) - A_{ij} \right] + \mathcal{O}(\delta_{ij}^2) \nonumber \\
&= \sum_{j=1}^N \left( -\frac{d_{ij}\delta_{ij}}{\epsilon} \right) A_{ij} + \mathcal{O}(\delta_{ij}^2)
\end{align}

Using \blauw{Eqn.\ \ref{eqn:Mijerror}} we now estimate the corresponding perturbations in the eigenvalues and eigenvectors. Expanding the perturbed eigenvalue problem in \blauw{Eqn.\ \ref{eqn:dmap_perturb}} and keeping terms to first order in the perturbations yields,
\begin{align}
&\mathbf{M} \mathbf{\Psi} + \mathbf{M} \mathbf{\delta\Psi} + \mathbf{\delta M} \mathbf{\Psi} = \mathbf{\Psi} \mathbf{\Lambda} + \mathbf{\Psi} \mathbf{\delta \Lambda} + \mathbf{\delta \Psi} \mathbf{\Lambda} \nonumber \\
\Rightarrow \; &\mathbf{M} \mathbf{\delta\Psi} + \mathbf{\delta M} \mathbf{\Psi} = \mathbf{\Psi} \mathbf{\delta \Lambda} + \mathbf{\delta \Psi} \mathbf{\Lambda}, \label{eqn:err6}
\end{align}
and where in going from the first to the second line we have canceled the first term on each side using original eigenvalue problem $\mathbf{M} \mathbf{\Psi} = \mathbf{\Psi} \mathbf{\Lambda}$ (\blauw{Eqn.\ \ref{eqn:dmap_dmap}}). Treating the unperturbed eigenvectors $\mathbf{\Psi}=\{\vec{\psi}_i\}_{i=1}^N$ as an orthonormal basis, we expand the perturbations to each eigenvector in this basis as,
\eqn{\vec{\delta\psi}_i = \sum_{j=1}^N \alpha_{j}^{(i)} \vec{\psi}_j,}{eqn:err7}
where $\{ \alpha_{j}^{(i)} \}_{j=1}^N$ are the expansion coefficients for the perturbation to the $i^\mathrm{th}$ eigenvector $\vec{\delta\psi}_i = \vec{\psi}_i^\prime - \vec{\psi}_i$. 

To solve for the perturbation to the $i^\mathrm{th}$ eigenvalue $\delta \lambda_i$ we restrict the perturbed eigenvalue problem in \blauw{Eqn.\ \ref{eqn:err6}} to the particular eigenvalue/eigenvector pair $\{ \lambda_i, \vec{\psi}_i \}$ and their corresponding perturbations $\{ \delta \lambda_i, \vec{\delta \psi}_i \}$ by extracting the $i^\mathrm{th}$ column, left multiplying each side by $\vec{\psi}_i^T$, and substituting in our expansion for $\vec{\delta\psi}_i$,
\begin{align}
&\vec{\psi}_i^T \left( \mathbf{M} \vec{\delta\psi}_i + \mathbf{\delta M} \vec{\psi}_i \right) = \vec{\psi}_i^T \left( \vec{\psi}_i \delta \lambda_i + \vec{\delta \psi}_i \lambda_i \right) \nonumber \\
\Rightarrow \; &\vec{\psi}_i^T \mathbf{M} \left( \sum_{j=1}^N \alpha_{j}^{(i)} \vec{\psi}_j \right) + \vec{\psi}_i^T \mathbf{\delta M} \vec{\psi}_i = \vec{\psi}_i^T \vec{\psi}_i \delta \lambda_i + \vec{\psi}_i^T \left( \sum_{j=1}^N \alpha_{j}^{(i)} \vec{\psi}_j \right) \lambda_i \nonumber \\
\Rightarrow \; &\sum_{j=1}^N \alpha_{j}^{(i)} \vec{\psi}_i^T \mathbf{M} \vec{\psi}_j + \vec{\psi}_i^T \mathbf{\delta M} \vec{\psi}_i = \delta \lambda_i + \lambda_i \sum_{j=1}^N \alpha_{j}^{(i)} \vec{\psi}_i^T \vec{\psi}_j \nonumber \\
\Rightarrow \; &\sum_{j=1}^N \alpha_{j}^{(i)} \lambda_j \vec{\psi}_i^T \vec{\psi}_j + \vec{\psi}_i^T \mathbf{\delta M} \vec{\psi}_i = \delta \lambda_i + \lambda_i \alpha_{i}^{(i)} \nonumber \\
\Rightarrow \; &\alpha_{i}^{(i)} \lambda_i + \vec{\psi}_i^T \mathbf{\delta M} \vec{\psi}_i = \delta \lambda_i + \lambda_i \alpha_{i}^{(i)} \nonumber \\
\Rightarrow \; &\delta \lambda_i = \vec{\psi}_i^T \mathbf{\delta M} \vec{\psi}_i \nonumber \\
&\quad \; \, = \vec{\psi}_i^T (\mathbf{D}^{-1} \mathbf{\delta A} - \mathbf{\delta D} \mathbf{D}^{-2} \mathbf{A}) \vec{\psi}_i \label{eqn:del_lambda}
\end{align}
where we have exploited the orthonormality of the eigenvectors, in going from the third line to the fourth used the original eigenvalue problem in \blauw{Eqn.\ \ref{eqn:dmap_dmap}} to make the substitution $\mathbf{M} \vec{\psi}_j = \lambda_j \vec{\psi}_j$, and used \blauw{Eqn.\ \ref{eqn:Merror}} to go from the penultimate to ultimate line.

Using a similar procedure we develop expressions for the expansion coefficients $\{ \alpha_{j}^{(i)} \}_{j=1}^N$ that we combine with \blauw{Eqn.\ \ref{eqn:err7}} to estimate perturbations in the eigenvectors. To compute $\alpha_{l}^{(i)}$ for $l \neq i$, we again extract the $i^\mathrm{th}$ column of the perturbed eigenvalue problem in \blauw{Eqn.\ \ref{eqn:err6}} but this time left multiply by $\vec{\psi}_l^T$,
\begin{align}
&\sum_{j=1}^N \alpha_{j}^{(i)} \vec{\psi}_l^T \mathbf{M} \vec{\psi}_j + \vec{\psi}_l^T \mathbf{\delta M} \vec{\psi}_i = \vec{\psi}_l^T \vec{\psi}_i \delta \lambda_i + \lambda_i \sum_{j=1}^N \alpha_{j}^{(i)} \vec{\psi}_l^T \vec{\psi}_j \nonumber \\
\Rightarrow \; &\sum_{j=1}^N \alpha_{j}^{(i)} \lambda_j \vec{\psi}_l^T \vec{\psi}_j + \vec{\psi}_l^T \mathbf{\delta M} \vec{\psi}_i = \lambda_i \sum_{j=1}^N \alpha_{j}^{(i)} \vec{\psi}_l^T \vec{\psi}_j \nonumber \\
\Rightarrow \; &\alpha_{l}^{(i)} = \frac{\vec{\psi}_l^T \mathbf{\delta M} \vec{\psi}_i}{\lambda_i - \lambda_l} \nonumber \\
&\quad \; \, =  \frac{\vec{\psi}_l^T (\mathbf{D}^{-1} \mathbf{\delta A} - \mathbf{\delta D} \mathbf{D}^{-2} \mathbf{A}) \vec{\psi}_i}{\lambda_i - \lambda_l}, \quad l \neq i. \label{eqn:alpha_il}
\end{align}
With the $\alpha_{l}^{(i)}$ for $l \neq i$ in hand, we recover $\alpha_{i}^{(i)}$ by enforcing normalization of the perturbed eigenvectors $\vec{\psi}_i^\prime = \vec{\psi}_i + \vec{\delta\psi}_i$,
\begin{align}
&\vec{\psi}_i^{\prime\;T} \vec{\psi}_i^\prime = 1 \nonumber \\
\Rightarrow \; &(\vec{\psi}_i + \vec{\delta\psi}_i)^T (\vec{\psi}_i + \vec{\delta\psi}_i) = 1 \nonumber \\
\Rightarrow \; &\vec{\psi}_i^T \vec{\psi}_i + 2 \vec{\psi}_i^T \vec{\delta\psi}_i + \vec{\delta\psi}_i^T \vec{\delta\psi}_i = 1 \nonumber \\
\Rightarrow \; &2 \vec{\psi}_i^T \left( \sum_{j=1}^N \alpha_{j}^{(i)} \vec{\psi}_j \right) + \left( \sum_{j=1}^N \alpha_{j}^{(i)} \vec{\psi}_j \right)^T \left( \sum_{l=1}^N \alpha_{l}^{(i)} \vec{\psi}_l \right) = 0 \nonumber \\
\Rightarrow \; &2 \sum_{j=1}^N \alpha_{j}^{(i)} \vec{\psi}_i^T \vec{\psi}_j + \sum_{j=1}^N \sum_{l=1}^N \alpha_{j}^{(i)} \alpha_{l}^{(i)} \vec{\psi}_j^T \vec{\psi}_l = 0 \nonumber \\
\Rightarrow \; &2 \alpha_{i}^{(i)} + \sum_{l=1}^N \left( \alpha_{l}^{(i)} \right)^2 = 0 \nonumber \\
\Rightarrow \; &\left( \alpha_{i}^{(i)} \right)^2 + 2 \alpha_{i}^{(i)} + \sum_{\substack{l=1 \\ l \neq i}}^N \left( \alpha_{l}^{(i)} \right)^2 = 0 \nonumber \\
\Rightarrow \; &\alpha_{i}^{(i)} = -1 + \sqrt{1 - \sum_{\substack{l=1 \\ l \neq i}}^N \left( \alpha_{l}^{(i)} \right)^2} \label{eqn:alpha_ii}
\end{align}
where we have appealed to the orthonormality of the eigenvectors $\{\vec{\psi}_i\}_{i=1}^N$ and in the last line solved the quadratic in $\alpha_{i}^{(i)}$ by completing the square and taking the positive root that corresponds to shrinking of the perturbed eigenvector along $\vec{\psi}_i$ to maintain normalization while preserving its original sense. Finally, we restrict our perturbative analysis (\blauw{Eqns.\ \ref{eqn:err7} and \ref{eqn:del_lambda}}) to the subspace spanned by the top $l = 2 \ldots (k+1)$ non-trivial eigenvalues $\{\lambda_i\}_{i=2}^{k+1}$ and eigenvectors $\{\vec{\psi}_i\}_{i=2}^{k+1}$ such that we model only the perturbations within the $k$-dimensional subspace containing the intrinsic manifold.

The result of our analysis is a first-order perturbative expression for the errors introduced by the landmarking procedure in the eigenvalues (\blauw{Eqn.\ \ref{eqn:del_lambda}}) and eigenvectors (\blauw{Eqns.\ \ref{eqn:err7}, \ref{eqn:alpha_il}, and \ref{eqn:alpha_ii}}) from our landmarking procedure.  In \blauw{Section \ref{sec:dmap_acc}} we demonstrate that for sufficiently small perturbations in the pairwise distances $\delta_{ij}$ (i.e., sufficiently many landmark points) that these perturbative analytical predictions are in good agreement with errors calculated by explicit numerical solutions of the full and landmark eigenvalue problems.

\subsection{Datasets}\label{sec:datasets}

\begin{figure*}[ht!]
\begin{center}
\includegraphics[width=16cm]{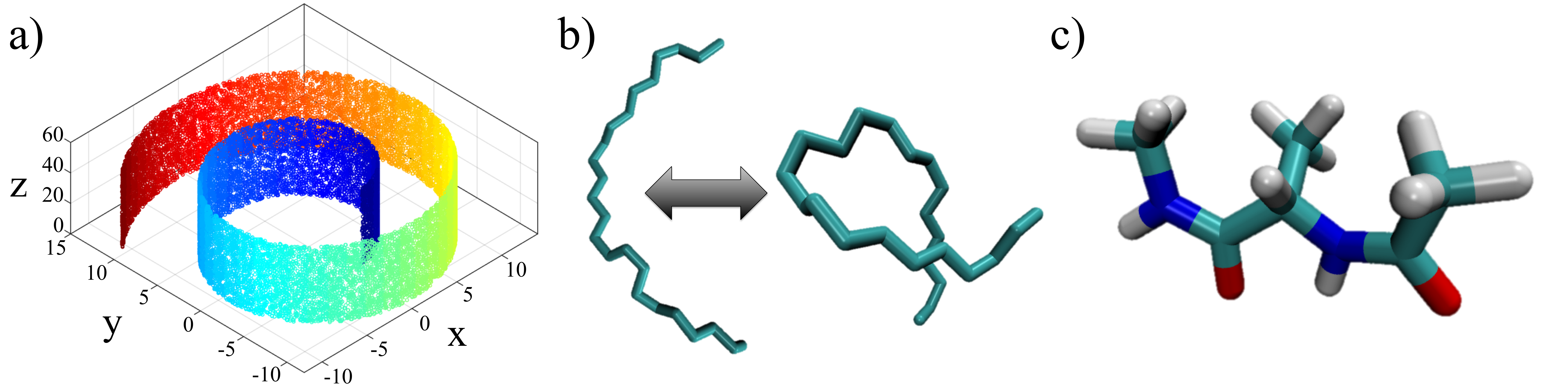}
\caption{The three systems to which L-dMaps was validated and benchmarked. (a) The Swiss roll, a 2D surface embedded in 3D space that is a canonical test system for nonlinear dimensionality reduction approaches\cite{Tenenbaum2000_Science}. (b) Molecular dynamics simulations of a \ce{C24H50} \textit{n}-alkane chain in water to which diffusion maps have previously been applied to discover hydrophobic collapse pathways\cite{Ferguson2010_PNAS,Wang2016_PhysRevE}. (c) Molecular dynamics simulations of alanine dipeptide -- the ``hydrogen atom of protein folding'' -- is the canonical test system for validating and benchmarking manifold learning, transition sampling, and metastable basin finding techniques in biomolecular simulation\cite{Ferguson2011_JChemPhys,Zheng2013_JPhysChemB,Hummer2003_JChemPhys,Chodera2006_MultiModSim,Ma2005_JPhysChemB,Stamati2010_Proteins,Michielssens2012_JPhysChemB,Chodera2007_JChemPhys}.}\label{fig:datasets}
\end{center}
\end{figure*}

We demonstrate and benchmark the proposed L-dMaps approach in applications to three datasets: the Swiss roll, molecular simulations of a \ce{C24H50} polymer chain, and biomolecular simulations of alanine dipeptide (\blauw{Fig.\ \ref{fig:datasets}}).

\textbf{Swiss roll.} The ``Swiss roll" -- a 2D surface rolled into 3D space -- is a canonical test system for nonlinear dimensionality reduction techniques \cite{Tenenbaum2000_Science,Ferguson2011_ChemPhysLett}. We illustrate in \blauw{Figure \ref{fig:datasets}a} the 20,000-point Swiss roll dataset employed by Tenenbaum \textit{et al.}\ in their introduction of the Isomap algorithm \cite{Tenenbaum2000_Science} and available for free public download from \url{http://isomap.stanford.edu/datasets.html}. In applying (landmark) diffusion maps to these data, distances between points are measured using Euclidean distances computed by MATLAB's L2-norm function.

\textbf{Molecular simulations of a \ce{C24H50} \textit{n}-alkane chain.} Hydrophobic polymer chains in water possess rich conformational dynamics, and serve as prototypical models for hydrophobic folding that we have previously studied using diffusion maps to determine collapse pathways\cite{Ferguson2010_PNAS} and reconstruct folding funnels from low-dimensional time series\cite{Wang2016_PhysRevE}. We consider a molecular dynamics trajectory of a \ce{C24H50} \textit{n}-alkane chain in water reported in \blauw{Ref.\ [\!\! \citenum{Wang2016_PhysRevE}]}. Simulations were conducted in the GROMACS 4.6 simulation suite\cite{GROMACS} at 298 K and 1 bar employing the TraPPE potential\cite{Martin1998_JPhysChemB} for the chain that treats each \ce{CH2} and \ce{CH3} group as a single united atom, and the SPC water model\cite{Berendsen1981_Springer}. This dataset comprises 10,000 chain configurations harvested over the course of a 100 ns simulation represented as 72-dimensional vectors corresponding to the Cartesian coordinates of the 24 united atoms. Distances between chains are computed as the rotationally and translationally minimized root-mean-square deviation (\groen{RMSD}), calculated using the GROMACS \emph{g\_rms} tool (\url{http://manual.gromacs.org/archive/4.6.3/online/g_rms.html}).

\textbf{Molecular simulations of alanine dipeptide.} Finally, we study a molecular dynamics simulation trajectory of alanine dipeptide in water previously reported in \blauw{Ref.\ [\!\!\citenum{Ferguson2011_JChemPhys}]}. The ``hydrogen atom of protein folding'', this peptide is a standard test system for new simulation and analysis methods in biomolecular simulation\cite{Ferguson2011_JChemPhys,Zheng2013_JPhysChemB,Hummer2003_JChemPhys,Chodera2006_MultiModSim,Ma2005_JPhysChemB,Stamati2010_Proteins,Michielssens2012_JPhysChemB,Chodera2007_JChemPhys}. Unbiased molecular dynamics simulations were conducted in the Gromacs 4.0.2 suite\cite{GROMACS} at 298 K and 1 bar modeling the peptide using the OPLS-AA/L force field\cite{Jorgensen1983_JChemPhys,Kaminski2001_JPhysChemB} and the TIP3P water model\cite{Jorgensen1988_JACS}. The dataset comprises 25,001 snapshots of the peptide collected over the course of a 50 ns simulation represented as 66-dimensional vectors comprising the Cartesian coordinates of the 22 atoms of the peptide. Distances between snapshots were again measured as the rotationally and translationally minimized RMSD calculated using the GROMACS \emph{g\_rms} tool.

\section{Results and Discussion}

The principal goal of L-dMaps is to accelerate out-of-sample extension by considering only a subset $M \ll N$ of landmark points at the expense of the fidelity of the nonlinear embedding relative to that which would have been achieved using all $N$ samples. First, we quantify the accuracy of L-dMaps nonlinear embeddings for both PST and k-medoid landmark selection strategies for the three datasets considered, and compare calculated errors with those estimated from our analytical expressions. Second, we benchmark the speedup and performance of L-dMaps for out-of-sample projection of new data.

\subsection{Landmark diffusion map accuracy}\label{sec:dmap_acc}

We numerically quantify the accuracy of L-dMaps manifold reconstruction for each of the three datasets using 5-fold cross validation, where we randomly split the data into five equal partitions and consider each partition in turn as the test set and the balance as the training set. We perform a full diffusion map embedding of the training partition to compute the ``true'' diffusion map embedding of the complete training partition $\mathbf{\Psi}_\textrm{train}$ using \blauw{Eqn.\ \ref{eqn:dmap_dmap}}. We then use the embedding of the training data onto the intrinsic manifold to perform the Nystr\"{o}m out-of-sample extension projection of the test data onto the manifold using \blauw{Eqn.\ \ref{eqn:nystrom}} to generate their ``true'' images $\mathbf{\Psi}_\textrm{test}$. 

We assess the fidelity of the nonlinear projections of the training data generated by our landmarking approach by taking the ensemble of training data and defining $M$ landmarks using the PST and k-medoids selection criteria. We then compute the nonlinear embedding of these landmarks onto the intrinsic manifold by solving the reduced eigenvalue problem in \blauw{Eqn.\ \ref{eqn:block_eig}} and projecting in the remainder of the training data (i.e., the $(N-M)$ non-landmark points) using the landmark Nystr\"{o}m extension in \blauw{Eqn.\ \ref{eqn:nystrom_block}}. This defines a landmark projection of the training data $\mathbf{\Psi}^\prime_\mathrm{train}$. Finally, we use the landmark Nystr\"{o}m extension again to project in the test partition to generate the landmark embedding of the test data $\mathbf{\Psi}^\prime_\mathrm{test}$.

To compare the fidelity of the landmark embedding we define a normalized percentage deviation between the true and landmark embeddings of each point as,
\eqn{\zeta(i) = 100\times\sqrt{\sum_{l=2}^{k+1} \left[ \frac{\vec{\psi}_l^\prime(i) - \vec{\psi}_l(i)}{\mathrm{range}(\vec{\psi}_l)} \right]^2},}{eqn:zeta_error}
where $\vec{\psi}_l(i)$ and $\vec{\psi}_l^\prime(i)$ are the true and landmark embeddings of data point $i$ into the $l^\mathrm{th}$ component of the nonlinear embedding into the $k$-dimensional intrinsic manifold spanned by the top $k$ non-trivial eigenvectors, $\mathrm{range}(\vec{\psi}_l) = \mathrm{max}(\vec{\psi}_l(i)) - \mathrm{min}(\vec{\psi}_l(i))$ is the linear span of dimension $l$, and the dimensionality $k$ is determined by a gap in the eigenvalue spectrum at eigenvalue $\lambda_{k+1}$. The dimensionality of the intrinsic manifolds for the Swiss roll, \ce{C24H50} chain, and alanine dipeptide have previously been determined to be 2, 4, and 2, respectively.\cite{Tenenbaum2000_Science,Wang2016_PhysRevE,Ferguson2011_JChemPhys}. We then compute the root mean squared (\groen{RMS}) normalized percentage error as,
\eqn{Z=\sqrt{\frac{1}{P}\sum_{i=1}^{P}\left(\zeta(i)\right)^2},}{eqn:Z_error}
where $P$ is the number of points constituting either the training or test partition.

\begin{figure*}[ht!]
\begin{center}
\includegraphics[width=16cm]{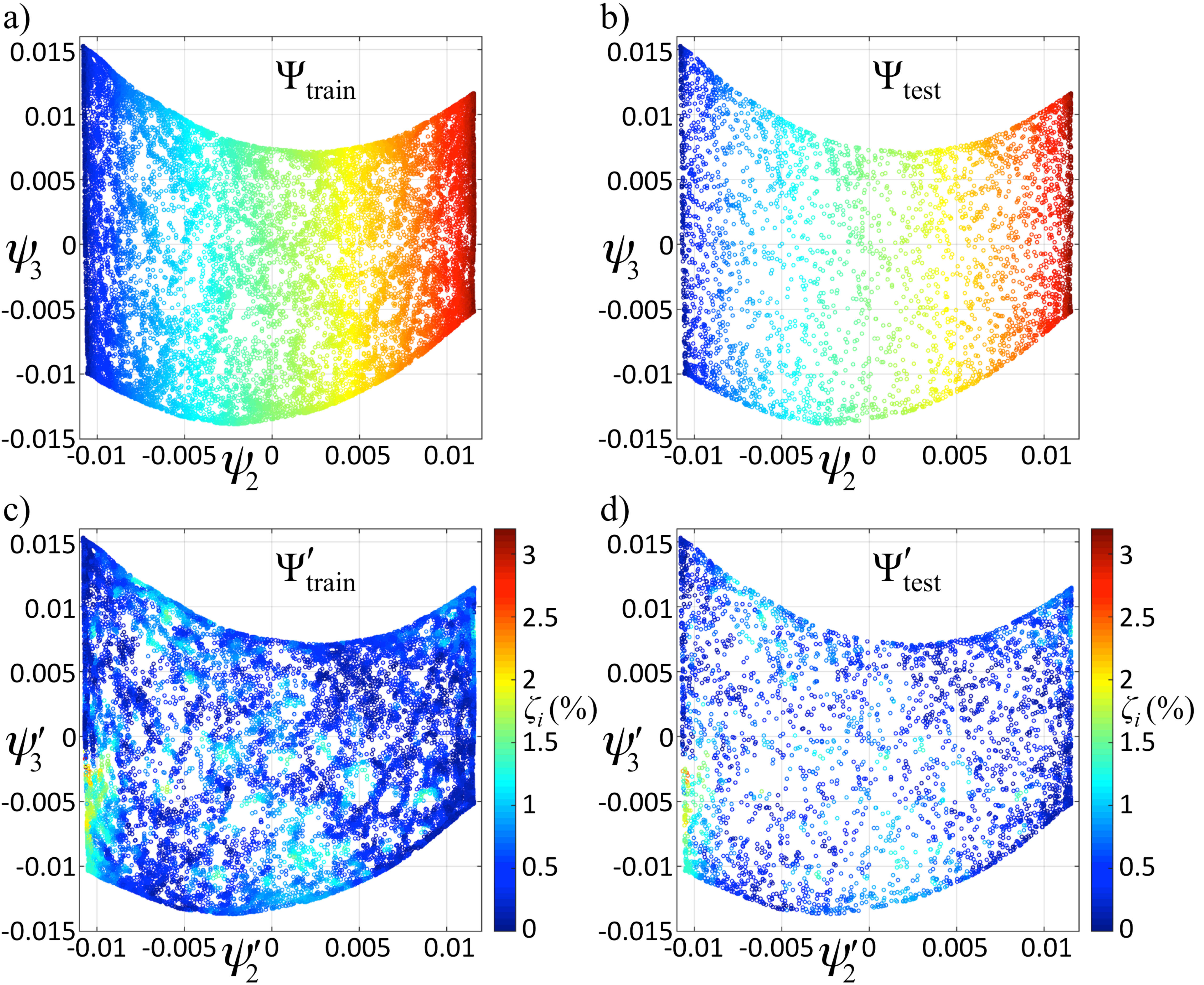}
\caption{Landmark error estimates for the $N$ = 20,000 point Swiss roll dataset split into a 80\% training ($N_\mathrm{train}$ = 16,000) and 20\% test ($N_\mathrm{test}$ = 4000) partitions. (a) Application of the original diffusion map to the full $N_\mathrm{train}$ = 16,000 training partition yields the 2D embedding $\mathbf{\Psi}_\mathrm{train} = \{ \vec{\psi}_{2,\mathrm{train}}, \vec{\psi}_{3,\mathrm{train}} \}$. (b) Nystr\"{o}m out-of-sample extension of the $N_\mathrm{test}$ = 4000 test points using the full embedding of the training data yields the 2D embedding $\mathbf{\Psi}_\mathrm{test} = \{ \vec{\psi}_{2,\mathrm{test}}, \vec{\psi}_{3,\mathrm{test}} \}$. (c) $M$ = 4,513 landmarks were selected from the $N_\mathrm{train}$ = 16,000 training points using the pruned spanning tree (PST) approach and used to define a landmark approximation to the intrinsic manifold into which the remaining $(N_\mathrm{train} - M)$ = 11,487 training points were embedded $\mathbf{\Psi}^\prime_\mathrm{train} = \{ \vec{\psi}^\prime_{2,\mathrm{train}}, \vec{\psi}^\prime_{3,\mathrm{train}} \}$. (d) Landmark Nystr\"{o}m projection of the $N_\mathrm{test}$ = 4000 test points into $\mathbf{\Psi}^\prime_\mathrm{test} = \{ \vec{\psi}^\prime_{2,\mathrm{test}}, \vec{\psi}^\prime_{3,\mathrm{test}} \}$. Points are colored in panels (a,b) by their position along the spiral of the roll as illustrated in \blauw{Fig.\ \ref{fig:datasets}a}, and in panels (c,d) by the normalized percentage deviation in the projection $\zeta(i)$ defined by \blauw{Eqn \ref{eqn:zeta_error}}.}\label{fig:swiss_roll}
\end{center}
\end{figure*}

We illustrate in \blauw{Figure \ref{fig:swiss_roll}} the true and landmark embeddings of the training $\{ \mathbf{\Psi}_\mathrm{train}, \mathbf{\Psi}^\prime_\mathrm{train} \}$ and test $\{ \mathbf{\Psi}_\mathrm{test}, \mathbf{\Psi}^\prime_\mathrm{test} \}$ data for the Swiss roll dataset, where we selected from the ensemble of $N_\mathrm{train}$ = 16,000 points in the training partition a set of $M$ = 4,513 landmarks using the PST algorithm. The maximum normalized percentage deviation of any one point in either the training or test data using this set of landmarks is less than 3.2\%, and the RMS errors are $Z_\textrm{train} = 0.643\%$ and $Z_\textrm{test} = 0.632\%$. This demonstrates that using only 28\% of the training set as landmarks, we can reconstruct a high-fidelity embedding with average normalized percentage errors per point of less than 1\%.

We present in \blauw{Table \ref{tab:errors}} the results of our error analysis for the three datasets using the PST and k-medoids landmark selection strategies at the smallest value of the kernel bandwidth $\epsilon$ supporting a fully connected random walk over the data (cf.\ \blauw{Eqn.\ \ref{eqn:dmap_kernel}}). This value of $\epsilon$ maximally preserves the fine-grained details of the manifold; we consider larger bandwidths below. For each of the three datasets we observe high-fidelity embeddings using relatively small fractions of the data as landmarks. For the Swiss roll, $\sim$30\% of the data are required to achieve a $\sim$2.5\% reconstruction error. Due to the approximately constant density of points over the manifold, further reductions provide too few points to support accurate embeddings. For the \ce{C24H50} \textit{n}-alkane chain and alanine dipeptide, the existence of an underlying energy potential produces large spatial variations in the density over the manifold, which is exploited by our density weighted landmark diffusion map (\blauw{Eqns.\ \ref{eqn:block_eig} and \ref{eqn:nystrom_block}}) to place landmarks approximately uniformly over the manifold and eliminate large numbers of points in the high density regions without substantial loss of accuracy. For \ce{C24H50}, PST landmarks constituting 1.19\% of the training data attain a RMS reconstruction error of $\sim$8\%, and k-medoids landmarks comprising 3.75\% of the data achieve a reconstruction error of $\sim$3\%. For alanine dipeptide, landmark selection using the PST selection policy achieves a $\sim$1\% error rate using only 1.75\% of the training points, whereas k-medoids requires more than 5\% of the data to achieve that same level of accuracy.

\begin{table}[ht!]
\centering
\begin{tabular}{||l||c|c|c|c||}
\hline
& \multicolumn{4}{c||}{\textbf{Swiss roll ($N_\textrm{train}=16000$, $N_\textrm{test}=4000$, $\epsilon=1$)}} \\
\hline
Method & \multicolumn{1}{c|}{$M$} & \multicolumn{1}{c|}{$M$ / $N_\textrm{train}$(\%)} & \multicolumn{1}{c|}{$Z_\textrm{train}$(\%)} & \multicolumn{1}{c||}{$Z_\textrm{test}$(\%) } \\
\hline
PST & 4551.0 (25.5) & 28.44 (0.16) & 2.42 (1.94) & 2.43 (1.95) \\
\hline
& 2000 & 12.50 & 13.43 (13.41) & 13.37 (13.30) \\
K-medoid & 4000 & 25.00 & 3.74 (3.15) & 3.75 (3.15) \\
& 8000 & 50.00 & 1.22 (1.05) & 1.22 (1.04) \\
\hline
& \multicolumn{4}{c||}{\textbf{\ce{C24H50} ($N_\textrm{train}=8000$, $N_\textrm{test}=2000$, $\epsilon=2.87\times 10^{-2}$)}} \\
\hline
Method & \multicolumn{1}{c|}{$M$} & \multicolumn{1}{c|}{$M$ / $N_\textrm{train}$(\%)} & \multicolumn{1}{c|}{$Z_\textrm{train}$(\%)} & \multicolumn{1}{c||}{$Z_\textrm{test}$(\%) } \\
\hline
PST & 95.2 (2.6) & 1.19 (0.03) & 7.85 (3.44) & 8.48 (3.56) \\
\hline
& 50 & 0.63 & 9.97 (2.86) & 10.65 (2.93) \\
K-medoid & 100 & 1.25 & 5.38 (1.23) & 5.76 (1.47) \\
& 300 & 3.75 & 3.19 (0.53) & 3.49 (0.65) \\
\hline
& \multicolumn{4}{c||}{\textbf{Alanine dipeptide ($N_\textrm{train}=20000$, $N_\textrm{test}=5000$, $\epsilon=1.06\times 10^{-3}$)}} \\
\hline
Method & \multicolumn{1}{c|}{$M$} & \multicolumn{1}{c|}{$M$ / $N_\textrm{train}$(\%)} & \multicolumn{1}{c|}{$Z_\textrm{train}$(\%)} & \multicolumn{1}{c||}{$Z_\textrm{test}$(\%) } \\
\hline
PST & 347.2 (8.4) & 1.74 (0.04) & 0.88 (0.26) & 0.94 (0.30) \\
\hline
& 200 & 1.00 & 5.93 (2.81) & 6.31 (3.04) \\
K-medoid & 400 & 2.00 & 2.92 (0.92) & 3.05 (0.84) \\
& 1000 & 5.00 & 1.43 (0.57) & 1.50 (0.60) \\
\hline
\end{tabular}
\caption{\textnormal{Root mean squared normalized percentage errors in the landmark nonlinear embeddings $Z$ for the PST and k-medoids landmark selection algorithms over the training and test partitions of the Swiss roll, \ce{C24H50}, and alanine dipeptide datasets. In each case the smallest value of $\epsilon$ supporting a fully connected random walk was employed in the kernel bandwidth. We report the mean and standard deviation of $Z_\textrm{train}$ and $Z_\textrm{test}$ estimated from 5-fold cross validation. For the PST landmark selection policy, we also report the mean and standard deviation in the number and percentage of landmark points.}}
\label{tab:errors}
\end{table}

\begin{figure*}[ht!]
\begin{center}
\includegraphics[width=16cm]{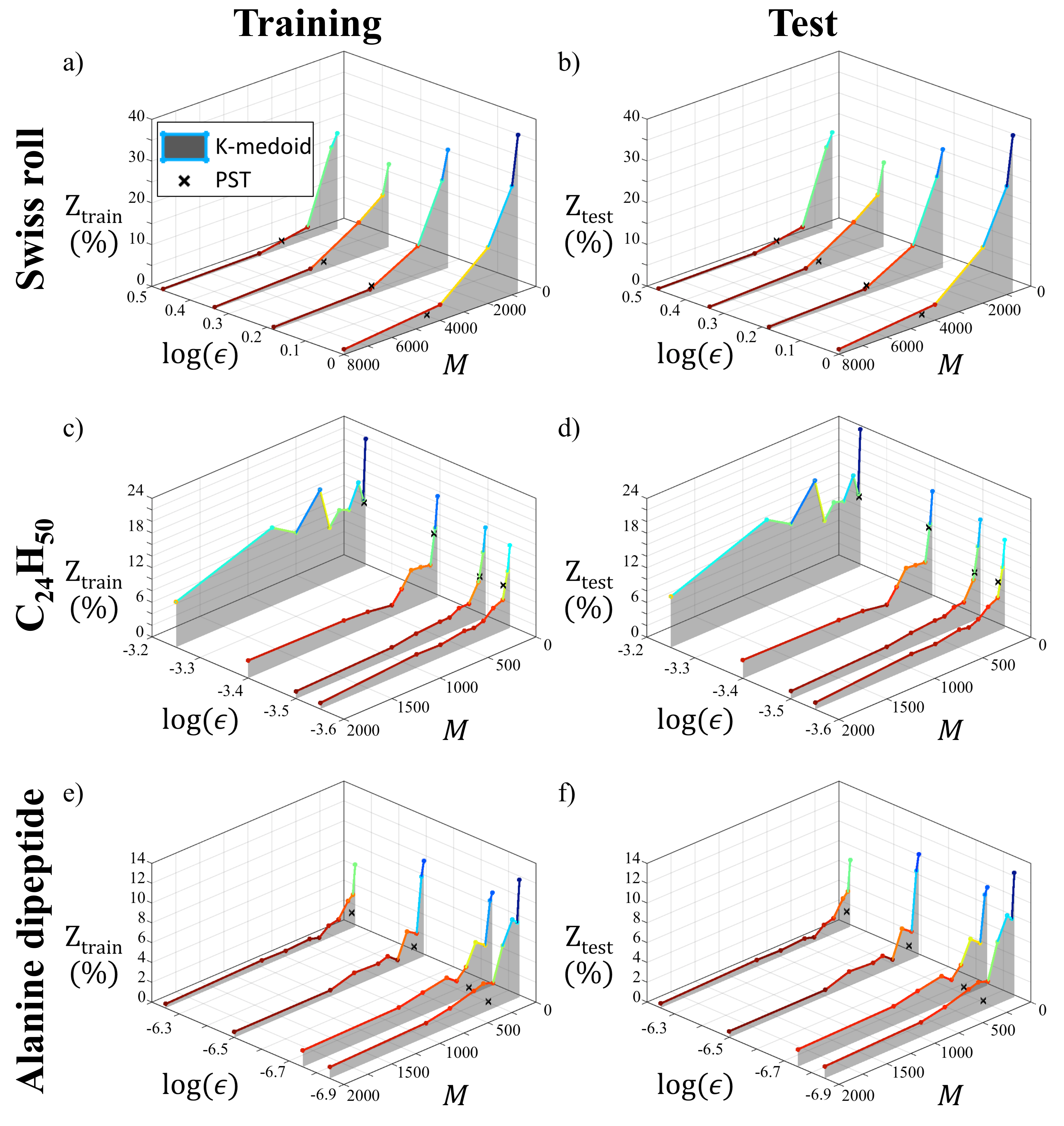}
\caption{Root mean squared normalized percentage errors in the landmark nonlinear embeddings $Z$ as a function of the PST (black crosses) or k-medoids (waterfall lines) landmark selection algorithm and kernel bandwidth $\epsilon$ for the training and test partitions of the (a,b) Swiss roll, (c,d) \ce{C24H50}, and (e,f) alanine dipeptide datasets. For the PST algorithm the number of landmarks $M$ is automatically selected as a function of $\epsilon$, whereas this is a user-defined adjustable parameter for the k-medoids approach. We plot the mean $Z_\textrm{train}$ and $Z_\textrm{test}$ estimated from 5-fold cross validation.}
\label{fig:waterfalls}
\end{center}
\end{figure*}

To further explore the relative performance of the PST and k-medoids landmark selection approaches on the three datasets we conducted a parametric analysis of the error rates for the two policies at a variety of kernel bandwidths $\epsilon$. Small $\epsilon$ values better resolve the fine-grained features of the manifold but require large numbers of landmark points to permit accurate interpolative embeddings of out-of-sample points by covering the manifold in overlapping $\sqrt{\epsilon}$ volumes. Large $\epsilon$ values sacrifice accurate representations of the details of the manifold, but permit the use of fewer landmark points. The PST selection procedure does not offer a means to directly tune the number of landmarks selected, which is controlled by $\epsilon$ used in construction of the spanning tree to ensure coverage and connectivity of the full dataset. The k-medoids policy permits the user to directly control the error to within a specified threshold by modulating the number of landmarks $M$. We present the results of our analysis in \blauw{Figure \ref{fig:waterfalls}}. The PST approach achieves better accuracy than k-medoids for the same number of landmarks, offering a good approach for automated landmark selection. The k-medoids error, however, can be tuned over a large range by adaptively choosing an appropriate number of landmark points. 

\begin{figure*}[ht!]
\begin{center}
\includegraphics[width=16cm]{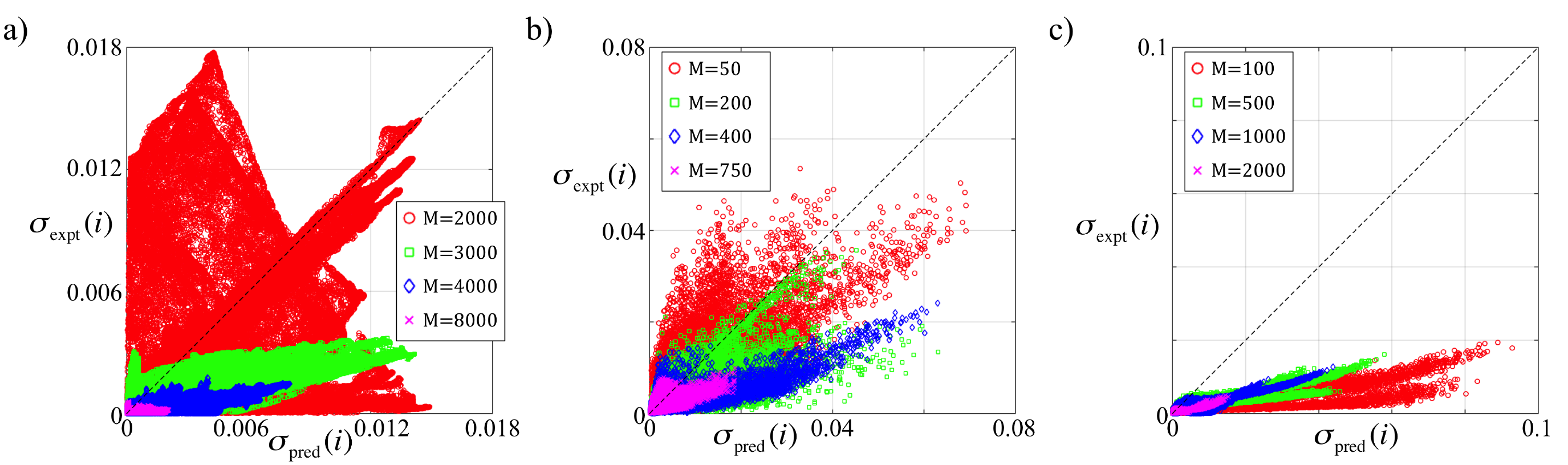}
\caption{Comparison of the predicted $\sigma_\mathrm{pred}(i)$ and calculated $\sigma_\mathrm{expt}(i)$ errors in the discrepancy of the embedding locations of the training data under the full and landmark embeddings for the (a) Swiss roll, (b) \ce{C24H50} \textit{n}-alkane chain, and (c) alanine dipeptide datasets.}
\label{fig:perturbation}
\end{center}
\end{figure*}

To make contact with our analytical error estimates developed in \blauw{Section \ref{sec:meth_err}}, we present in \blauw{Figure \ref{fig:perturbation}} a parity plot of the discrepancies in the embedding of the $i = 1,\ldots,N_\mathrm{train}$ training points between the full and landmark embeddings,
\begin{align}
\sigma(i) &= || [ \vec{\psi}_{2,\mathrm{train}}^\prime(i),\vec{\psi}_{3,\mathrm{train}}^\prime(i),\ldots,\vec{\psi}_{k+1,\mathrm{train}}^\prime(i) ] - [ \vec{\psi}_{2,\mathrm{train}}(i),\vec{\psi}_{3,\mathrm{train}}(i),\ldots,\vec{\psi}_{k+1,\mathrm{train}}(i) ] || \nonumber \\
&= || [ \vec{\delta\psi}_{2,\mathrm{train}}(i),\vec{\delta\psi}_{3,\mathrm{train}}(i),\ldots,\vec{\delta\psi}_{k+1,\mathrm{train}}(i) ] || \nonumber \\
&= \sqrt{ \sum_{l=2}^{k+1} \left( \vec{\delta\psi}_{l,\mathrm{train}}(i) \right)^2 },
\end{align}
predicted from our analytical first-order perturbative expressions $\sigma_\mathrm{pred}(i)$ (\blauw{Eqns.\ \ref{eqn:err7}, \ref{eqn:alpha_il}, and \ref{eqn:alpha_ii}}) and calculated directly from our numerical computations $\sigma_\mathrm{expt}(i)$. We collate data from the k-medoids landmark selection process at all numbers of landmarks $M$ with the $\epsilon$ bandwidths provided in \blauw{Table \ref{tab:errors}}; perfect prediction would correspond to all points lying along the diagonal. The experimental and predicted errors show quite good agreement for large values of $M$ where the first-order perturbation expansion is a good approximation and the number of landmarks $M$ is such that the characteristic distance between them is on the order of $\sqrt{\epsilon}$. The agreement deteriorates for small numbers of landmarks where there are too few supports covering the manifold to admit an accurate perturbative treatment. In the large-$M$ / small-error regime where the perturbative treatment is accurate, the analytical error expressions can be used to predictively identify landmark selections to achieve a specified error tolerance.

\subsection{Out-of-sample extension speedup}

\begin{figure*}[ht!]
\begin{center}
\includegraphics[width=16cm]{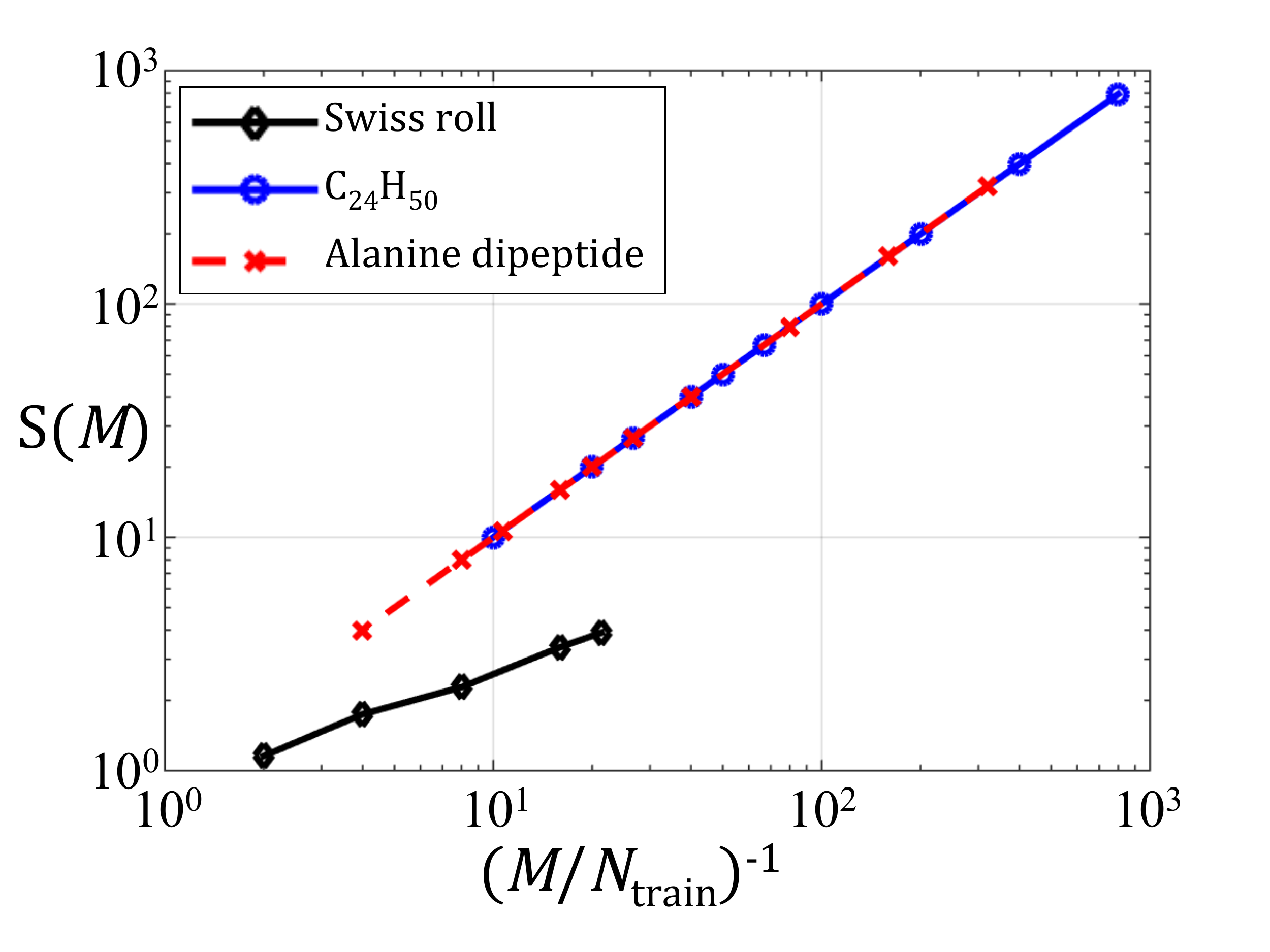}
\caption{Speedup $S(M) = t_\mathrm{full}/t_\mathrm{landmark}(M)$ in the embedding of the $N_\mathrm{test}$ test points using a landmark Nystr\"{o}m embedding over $M$ landmark points relative to a full embedding over all $N_\mathrm{train}$ points for each of the three datasets. Timings include both the computation of pairwise distances to the out-of-sample point and the Nystr\"{o}m projection procedure.}\label{fig:speedup}
\end{center}
\end{figure*}

In \blauw{Figure \ref{fig:speedup}} we present as a function of the number of landmarks $M$ the measured speedups $S(M) = t_\mathrm{full}/t_\mathrm{landmark}(M)$ for the k-medoids landmark selection data presented in \blauw{Figure \ref{fig:waterfalls}} at the $\epsilon$ values given in \blauw{Table \ref{tab:errors}}. $t_\mathrm{full}$ is the measured execution time for the Nystr\"{o}m embedding of the $N_\mathrm{test}$ training points using the locations of all $N_\mathrm{train}$ training points computed using the full diffusion map, and $t_\mathrm{landmark}(M)$ is the runtime for the landmark Nystr\"{o}m embedding using $M < N_\mathrm{train}$ landmark points. All calculations were performed on an Intel i7-3930 3.2GHz PC with 32GB of RAM, with landmark selection and diffusion mapping performed in MATLAB and distances computed as described in \blauw{Section \ref{sec:datasets}}.

For each of the three systems studied, we observe large accelerations in runtime as we decrease the number of landmarks employed. In the case of \ce{C24H50} and alanine dipeptide, we observe excellent agreement with the theoretically predicted $S \propto N_\mathrm{train}/M$. Selecting $M/N_\mathrm{train}\approx 4\%$ for \ce{C24H50} and $M/N_\mathrm{train}\approx 2\%$ for alanine dipeptide, we achieve 25-fold and 50-fold accelerations in the out-of-sample embedding runtime, respectively, while incurring only $\sim$3\% errors in the embedding accuracy (\blauw{Table \ref{tab:errors}}). Reduced accelerations and deviation from the expected scaling relation are observed for the Swiss roll dataset due to the very low cost of the Euclidean pairwise distance computation, which leaves these calculations dominated by computational overhead as opposed to the computational load of the pairwise distance and Nystr\"{o}m calculations. Nevertheless, we still achieve a two-fold acceleration in runtime for $M/N_\mathrm{train} \approx 25\%$ with a $\sim$4\% embedding error (\blauw{Table \ref{tab:errors}}). Using landmark diffusion maps can thus drastically reduce the out-of-sample restriction time, with particular efficacy in embedding samples with computationally expensive distance measures.

\section{Conclusions}

We have introduced a new technique to accelerate the nonlinear embedding of out-of-sample data into low-dimensional manifolds discovered by diffusion maps using an approach based on the identification of landmark points over which to perform the embedding. In analogy with the landmark Isomap (L-Isomap) algorithm \cite{Tenenbaum2000_Science,deSilva2002_NIPS}, we term our approach landmark diffusion maps (L-dMaps). By identifying with the $N$ data points a small number of $M \ll N$ landmarks to support the calculation of out-of-sample embeddings, we massively reduce the computational cost associated with this operation to achieve theoretical speedups $S \propto N/M$. We have validated the accuracy and benchmarked the performance of L-dMaps against three datasets: the Swiss roll, molecular simulations of a \ce{C24H50} polymer chain, and biomolecular simulations of alanine dipeptide. These numerical tests demonstrated the capacity of our approach to achieve up to 50-fold speedups in out-of-sample embeddings with less than 4\% errors in the embedding fidelity for molecular systems. L-dMaps enables the use of diffusion maps for rapid online embedding of high-volume and high-velocity streaming data, and is expected to be particularly valuable in applications where runtime performance is critical or rapid projection of new data points is paramount, such as in threat detection, anomaly recognition, and high-throughput online monitoring or analysis.

We observe that further accelerations to L-dMaps may be achieved using efficient algorithms to eliminate the need to compute all $M$ pairwise distances between the new out-of-sample point and the landmarks. For example the recently proposed Fast Library for Approximate Nearest Neighbors (FLANN)\cite{Muja2014_TPAMI} uses a tree-search to perform a radial k-nearest neighbors search, and has been previously employed in the construction of sparse diffusion maps\cite{McQueen2016_JMachLearnRes}. These techniques may also prove valuable in eliminating the need to compute the $N$-by-$N$ pairwise distances matrix required for landmark identification, the storage and analysis of which can prove prohibitive for large datasets. Techniques such as FLANN provides a means to efficiently construct sparse approximations to the kernel matrix $\mathbf{A}$ in which small hopping probabilities associated with large distances below a user-defined tolerance are thresholded to zero. Accordingly, we foresee combining approximate nearest neighbor distance computations, sparse matrix representations, and landmark embedding procedures as a fruitful area for future development of fast and efficient out-of-sample nonlinear embedding approaches.

\begin{acknowledgement}
Funding: This material is based upon work supported by a National Science Foundation CAREER Award to A. L. F. (Grant No. DMR-1350008). The sponsor had no role in study design, collection, analysis and interpretation of data, writing of the report, or decision to submit the article for publication
\end{acknowledgement}



\clearpage
\newpage

\bibliography{manuscript}

\end{document}